%% file: main.tex
\documentclass{svproc}
\usepackage{url}

\usepackage{times}
\usepackage{latexsym}
\usepackage{graphicx}
\usepackage{amsmath}
\usepackage{amssymb}
\usepackage{multirow}
\usepackage[font=small,labelfont=bf,tableposition=top]{caption}
\usepackage[ruled]{algorithm}
\usepackage[]{algpseudocode}
\usepackage{enumitem}
\usepackage{subcaption}

\usepackage{colortbl}
\usepackage{xcolor}
\usepackage{booktabs}

\usepackage{microtype}
\begin{document}
\mainmatter              
\title{``How to best say it?'' : Translating Directives in Machine Language into Natural Language in the Blocks World}
\titlerunning{How to best say it?}  
%


\author{Sujeong Kim \and Amir Tamrakar}
\authorrunning{Kim and Tamrakar} 
%
\tocauthor{Sujeong Kim and Amir Tamrakar}
\institute{SRI International, Princeton NJ, USA,\\
\email{\{sujeong.kim|amir.tamrakar\}@sri.com}}

\maketitle              

\begin{abstract}
We propose a method to generate \emph{optimal} natural language for block placement directives generated by a machine's planner during human-agent interactions in the blocks world. A non user-friendly machine directive,  \emph{e.g., move(ObjId, toPos)}, is transformed into visually and contextually grounded referring expressions that are much easier for the user to comprehend. We describe an algorithm that progressively and generatively transforms the machine's directive in ECI (Elementary Composable Ideas)-space, generating many alternative versions of the directive. We then define a cost function to evaluate the \emph{ease of comprehension} of these alternatives and select the best option. The parameters for this cost function were derived empirically from a user study that measured \emph{utterance-to-action} timings. 
\end{abstract}

\input{1_introduction}

\input{2_background}

\input{2_method}
\input{3_userstudy}

\input{4_conclusion}

%
%
\bibliographystyle{spmpsci_unsrt}  
\bibliography{blocksworldref.bib}

\end{document}

%% file: 1_introduction.tex
\section{Introduction}






AI agents interacting with humans on collaborative tasks need to be able to communicate with the humans. Effective communication requires not only the ability to understand natural language, but also the ability to generate natural language that their human partners can comprehend with ease and accuracy. 
In this paper, we discuss the problem of generating optimal natural language utterances for human-machine interactions in the context of the Blocks World{~\cite{SHRDLU-WINOGRAD19721}}, specifically, for machine-to-human directives for block manipulation.




The machine level representations of such directives tend to be very precise but mechanistic, often resulting from mathematical operations or dictated by the needs of robotic manipulation interfaces. These directives, in their raw forms, are not at all intuitive for humans. To communicate them effectively, the agent needs to transform them into visually and contextually grounded descriptions that map closely to how humans perceive and describe the world~\cite{Grounded-Semantic-Composition-Gorniak-2004,SimulatedPG-Thorisson-1994,visual-salience-and-perceptual-grouping-landragin-2001,multimodal-referring-expressions-krahmer-van-der-sluis-2003,referring-expressions-dale-and-reiter-1995,logic-and-conversation-grice-1975}.

We introduce the concept of a generative transformation that recursively constructs more complex descriptions of objects and/or locations in relation to other objects or properties of objects, i.e. generates referring expressions, using the context of the scene and/or the discourse context. We enforce \emph{unambiguousness} and strive for \emph{conciseness} in the generated forms. Even with these constraints, the process generates many alternative forms for the directive. We pick the \emph{optimal} utterance from this set by evaluating its \emph{efficiency} based on the ease of the recipient to process the language and act on it. We evaluate this based on a novel \emph{empirically} determined cost function that computes weighted description lengths of each utterance. The empirical weights were obtained from human user studies that measured the time taken for comprehending the directive (and successfully performing the action prescribed by the directive) and factoring out the contributions of the different types of properties (Section \ref{sec:user_study}).

%% file: 2_background.tex
\begin{figure}[t]
\centering
\includegraphics[width=\textwidth]{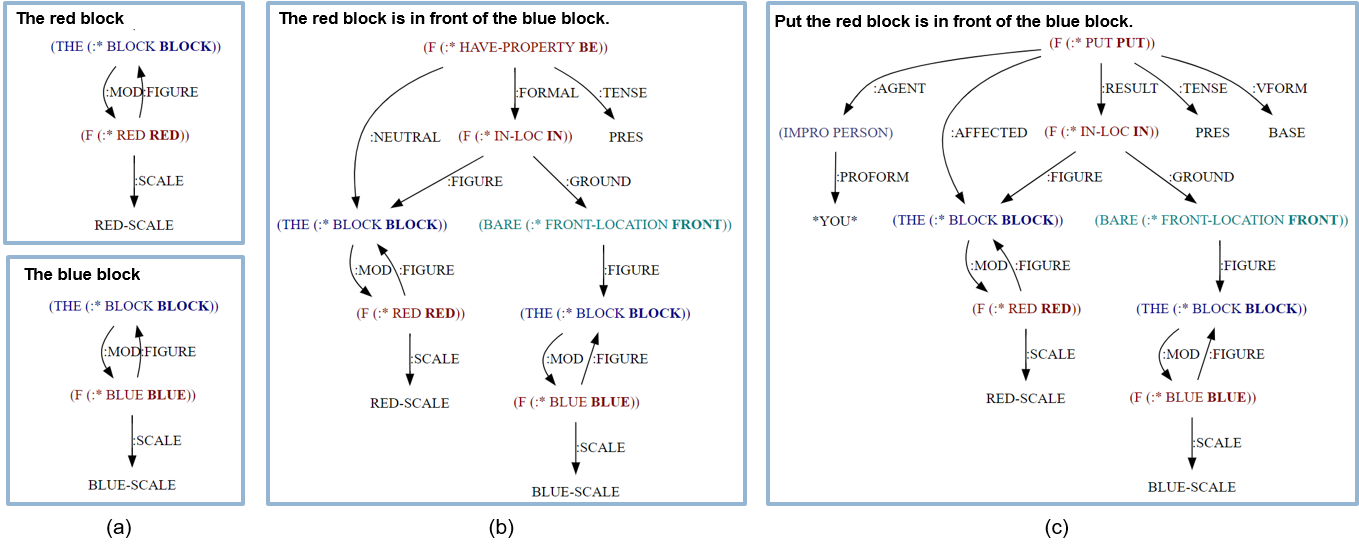}
\caption{ Examples showing basic ECIs and more complex ECIs composed with these basic ECIs. (a) two ECIs, each representing an object with their color property, (b) an ECI describing spatial relationship between the two ECIs, and (c) a directive ECI composed with the ECIs from (a) and (b). The SVG diagrams are generated using the TRIPS cogent parser web interface ~\cite{TripsParserWebInterface}.}
\label{fig:EGT_object_description_example}
\end{figure} 

\section{Background and Related Work}

Effective communication between human and machine is one of the important factors for robust human-agent interactions~\cite{kozierok2021hallmarks}. For a collaborative tasks in the Blocks World, communications tend to be task-oriented, requiring description of the objects or manipulation of them. We approach this problem by generating grounded language, similar to how people verbally describe objects in the scene~\cite{Grounded-Semantic-Composition-Gorniak-2004}, and based on the understanding of human perception.

In this route, there have been efforts to generate a perceived model of the scene. Landragin et al. discuss a saliency model based on visual familiarity, intentionality, and physical characteristics~\cite{visual-salience-and-perceptual-grouping-landragin-2001}, Th\'{o}rrisson also discuss proximity and similarity factors of perceiving objects as a group~\cite{SimulatedPG-Thorisson-1994}. Generating referring expressions of an entity based on their properties is also an important problem and has been studied by multiple scholars~\cite{referring-expressions-dale-and-reiter-1995,multimodal-referring-expressions-krahmer-van-der-sluis-2003}. Last but not least, evaluating and optimizing referring expressions or machine-generated instructions allows us to generate more effective dialogues. For example, one can use principles like Gricean maxims of conversation~\cite{logic-and-conversation-grice-1975} or conduct studies to explore different strategies~\cite{Foster2009}.

We utilize the representation of Elementary Composable Ideas (ECIs) introduced in the DARPA-CwC program to formulate our generative transformation algorithm to translate machine-level directives into human-comprehensible directives. ECIs are minimal basic concepts that can be composed into Composite ECIs (or CCIs) to represent complex ideas. There have been efforts to identify sets of ECIs and define their representations~\cite{ECIpedia:McDonald:2018,trips-allen-2017}, application of ECIs in the music composition domain~\cite{Donya2017MusECI}.

The ECIs and CCIs can be readily converted into English language sentences, and vice versa. Figure~\ref{fig:EGT_object_description_example} shows example ECIs, from the ones representing basic concepts, such as a single object with its property, to more complex ideas composed with multiple simpler ECIs.


Our method, ECI Generative Transformation (EGT), is a novel algorithm to compose ECIs using grounding and optimization. We adopt prior approaches by~\cite{SimulatedPG-Thorisson-1994,visual-salience-and-perceptual-grouping-landragin-2001} in utilizing visually salient properties arising out of perceptual grouping and ~\cite{multimodal-referring-expressions-krahmer-van-der-sluis-2003,referring-expressions-dale-and-reiter-1995} for describing objects using their properties. In addition, we introduce use of discourse and task context whenever possible. By utilizing ECI structures, our method provides an efficient and structural way in generating grounded language.

%% file: 2_method.tex
\section{Grounded Natural Language Generation}\label{sec:perception_modules}


\begin{figure}[t]
\centering
\includegraphics[width=1.0\textwidth]{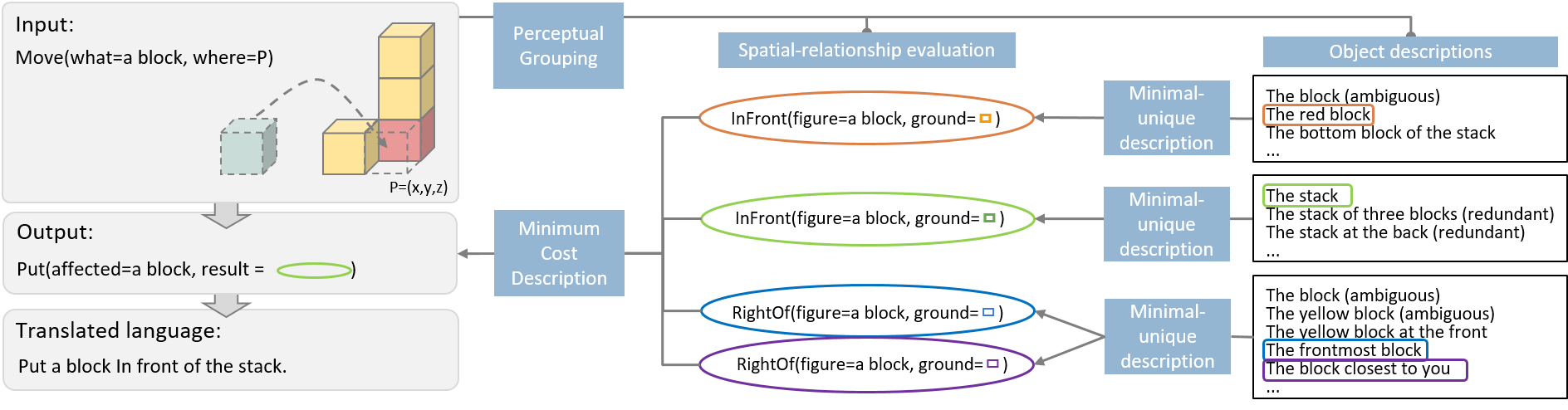}
\caption{\textbf{Overview of the EGT process}: A block-movement plan is converted to a \textit{Put-ECI} that later gets translated into a natural langauge. The rectangles highlighted in blue are the main components of the EGT. Empty colored boxes or ovals correspond to the ECIs highlighted in the same colored shape.}
\label{fig:overview}
\end{figure}



\paragraph{Problem Definition}

Assume the agent wants the human to move a block, $b$, to a new position. The machine's low-level planner will generate directives like \emph{move(objId=$b$, toPos=$\mathbf{p}=(x',y',z'))$} or \emph{translate(objId=$b$, dir=$(\Delta x, \Delta y, \Delta z))$}. Our goal is to transform this machine language into human understandable natural language. We also assume that the agent is privy to a full scene description to use for grounding, as well as the full history of past interactions to use as discourse context.

\subsection{ECI Generative Transformation (EGT)}\label{sec:EGT}



Figure \ref{fig:overview} shows an overview of our approach, which we call \emph{ECI Generative Transformation} (EGT) because the transformations are essentially operations that splice new ECIs into the existing complex idea (CCIs) representing the directive, thereby creating a new form of it. The main idea of EGT is to consider all possible reference objects in the scene and attempt to generate descriptions of the target location, in the machine-directive, using all spatial relationship concepts available in the ontology relative to these reference objects. ECIs capture these relationships. Only compositions of these relationships (using the AND-ECI) that can uniquely identify the location are kept. Furthermore, when the reference objects used in these ECIs cannot themselves be uniquely described based on their available property-ECIs, we recurse to the next depth of the algorithm where we try to uniquely identify these reference objects based on their spatial relationship with some other reference objects in the world, only giving up if we get to a predefined maximum recursion depth. Algorithm~\ref{alg:main} describes this process in more detail. The next few subsections describe various operations described in the algorithm listing.

\begin{algorithm}[h]
        \footnotesize
		\caption{Converts a \emph{move(b, $p$)} into a Put-ECI directive.}
		\label{alg:main}
\begin{algorithmic}[1]
    \Require $b'$: imaginary block identical to $b$, $\mathbf{p}$: target position, $\mathbf{U}$: set of all blocks in the scene, $d_{max}$: maximum depth of recursion
    
	\Procedure{MachineInstructionToECI}{move($b$, $\mathbf{p})$}
	\State $\mathbf{CS} :=$ PerceptualGrouping($\mathbf{U})\cup\mathbf{U}$
	
	\State $b'.pos := \mathbf{p}$
	
	\State $\mathbf{D} := $DescribeUniquely$(b', \mathbf{CS})$
	
	\State $\mathbf{D'}$ = ECify(Put, affected=$b$, result = $\mathbf{D}$)
	\State $\mathbf{D''}$ = ECify(ActionContext($\mathbf{D'}$))
	
	\State return $\{d\in \mathbf{D'}\cup \mathbf{D''}| $C($d) = $min(C$(\mathbf{D'}\cup \mathbf{D''}))$\}
	\EndProcedure

\Procedure{DescribeUniquely}{$b$, $\mathbf{CS}$}
\If{$d > d_{max}$}
  return $\emptyset$
\EndIf

\State $\mathbf{N}:=\{(g,r)|\exists$SpatialReln(grnd=$g$,figr=$b$,type=$r$)\}
\For{$(g, r) \in \mathbf{N}$}
    \State $\mathbf{D} := $MinimalUniqueDescriptions$(g)$
    \If{$\mathbf{D}\neq\emptyset$}
        \State return ECIfy(r, D, b)
    \Else
        \State $\mathbf{CS'} \gets \{ u | u \in \mathbf{CS}, u \neq g \}$
        \State $\mathbf{D'}:=$DescribeUniquely($g, \mathbf{CS'}$)
        \State return ECIfy($r, \mathbf{D'}, b$)
\EndIf
\EndFor
\EndProcedure
\end{algorithmic}
\end{algorithm}







\vspace{-0.1in}
\subsection{Perceptual Grouping}





According to Gestalt principles of visual perception, objects are perceived as a group under certain conditions. We incorporate and extend \cite{SimulatedPG-Thorisson-1994,visual-salience-and-perceptual-grouping-landragin-2001} for our blocks-world environment. Our Blocks World is a continuous (non-grid) environment: blocks do not perfectly align with each other. We detect formations based on \textit{proximity} and \textit{continuity}. Examples of such groupings are linear formations of blocks, \emph{e.g.} rows, columns/stacks. Detected formations are further subdivided into smaller subsets based on the visual properties, \emph{e.g.} colors. 





\subsection{Spatial Relationship Evaluation}\label{sec:spatialRelationship}


We utilize various predicates that represent spatial relationships between  objects. These include basic directional predicates like top, below, front, back, left, right, composite predicates like next-to, near, far, and location specific predicates like at-corner, at-center, etc.


When a predicate is evaluated to be true, we generate an ECI corresponding to its relationships in the form of \emph{spatialRelation-ECI}$(\textit{figure}, \textit{ground})$. For example, an ECI describing object $x$ in front of object $y$ is \emph{inFront}$(\textit{figure}=x, \textit{ground}=y)$. The set of viable \emph{spatialRelation}-ECIs are saved for use during the EGT iterations.


\subsection{Minimal Unique Description}\label{sec:referringExpression}

Another important component of the EGT algorithm is generating referring expressions for the objects. We define a set of salient properties to use as descriptors of an object (see Table~\ref{tab:descriptors}). 

We utilize two criteria: \textit{conciseness} and \textit{unambiguousness} in creating referring expressions. Conciseness is a criterion about the complexity and/or length of the language. The language should contain necessary information to be accurately parsed and interpreted without redundancy. Unambiguousness is a criterion about the accuracy of the language~\cite{logic-and-conversation-grice-1975}. For example, \emph{``the blue block at the top''} contains redundant information when there is only one blue block. In this case, \emph{``the blue block''} is the concise form of the referring expression. On the other hand, \emph{``the blue block''} is ambiguous when there are multiple blue blocks. Our goal is thus to generate unambiguous descriptions of an object while being as concise as possible. 

Let us denote $\mathbf{SD}_{i}$ a set of all combinations of descriptors for object $i$. 
A set of unique descriptors $\mathbf{D}_{i}$ is generated by choosing non-overlapping descriptors to all other objects.
\vspace{-1em}
\begin{equation}
   \mathbf{D}_i =  \{ s | s\in{\mathbf{SD}_i}, s\notin \bigcup\mathbf{SD}_j, j\neq i\}.
\end{equation}
Finally, we take descriptors with minimum-cost:
\vspace{-1em}
\begin{equation}
\mathbf{D}'_i =  \{ \mathbf{s} | \mathbf{s}\in{\mathbf{D}_i}, C(s)=min(C(\mathbf{D}_{ik})),
\end{equation}
where $k$ is a descriptor combination and $C$ is the cost function. 
We use empirically determined weighted costs for descriptors as described in Section \ref{sec:cost}. As result, we generate set of ECIs each describing an object uniquely using minimum number of properties.

\begin{table}[t]
\footnotesize \centering
\begin{tabular}{lll}
\toprule
 Acronym & Property & Examples  \\ \midrule
 REF& type &  block, row, stack, column\\
 COL & color & red, yellow, blue, ...  \\
 DM & degree modifier & frontmost, leftmost, ...\\
 CON & context & it, previous block, ...\\
 CNT & count & one, two, ...\\
 ORD & order & first, second, ...\\
 LOC & locality & top, bottom, middle, ...
  \\\bottomrule
\end{tabular}
\caption{\label{tab:descriptors} Categories and examples of properties used to generate referring expressions of an object.}
\end{table}

\subsection{Injecting Context}\label{sec:context}

We additionally generate directives based on the dialog-, task-, and action-context. \textit{Dialog-context} is used for the previously mentioned block. We replace its description with ``it". \textit{Task-context} is used for the previous block acted upon. We replace its object description with ``it", ``the previous block", ``the block you just put/placed/moved". \textit{Action-context} is used when current action is repetition of the previous action. We simply say ``add/put one more'', ``repeat one more time''. This process corresponds to line 6 of Algorithm~\ref{alg:main}.

\subsection{Evaluating Efficiency: The Cost Function}
\label{sec:cost}


We assume that the performance is inversely proportional to the complexity of the language.
We define performance as time taken to respond with an action after listening to the language. Then, the cost, $C$, can be found from the mapping between response time and the complexity of the language:
\begin{equation}
    C(\mathbf{D}) = \mathbf{C}_p^d\mathbf{D}_l+b,
    \label{eq:cost}
\end{equation}
where $\mathbf{D}_l$ is a feature vector representing the complexity of a language $l$, $\mathbf{C}_p^d$ is the coefficient matrix corresponding to the cost of each feature at different depth levels. 

%% file: 3_userstudy.tex
\section{User Study to Estimate Cost Function Parameters} \label{sec:user_study}

\begin{figure}[h]
\centering
\includegraphics[width=0.5\textwidth]{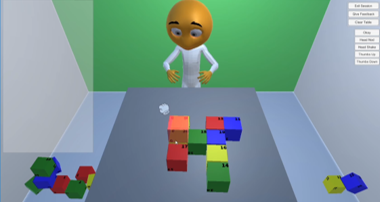}
\caption{SMILEE blocks-world system used for our experiment.}
\label{fig:virtual_setup}
\end{figure}

We conducted an in-depth user study to estimate the cost of the directives. We generated 543 test directives from 66 different block-placement scenarios using the EGT algorithm described in Section~\ref{sec:EGT}. The directives varied in depth, number and type of descriptors used. We used the publicly available system called SMILEE~\cite{kim-smilee-2018}, shown in Fig.~\ref{fig:virtual_setup} to perform the experiment. The natural language directives were passed through a Text-To-Speech engine and presented in audio form.  
Subjects performed block-placement tasks as directed by the agent. Each scenario included three sequential directives to generate some local historical context. We measured response time from the moment the agent started speaking to the moment where the subject completed their action. We also recorded the accuracy of the subject's actions. 
Table~\ref{tab:perceptual_study_stats} shows the summary of result. Depth-0 languages include directives generated from action context described in section~\ref{sec:context}. As depth increases, response time increases. High performance accuracies at all levels guarantees that the  response time reflects the time taken to fully understand the language.

\begin{table}[h]
\footnotesize \centering
\resizebox{0.48\textwidth}{!}{%
\begin{tabular}{@{}crrrr@{}}
depth & \multicolumn{1}{c}{\#properties} & \multicolumn{1}{c}{resp. time} & \multicolumn{1}{c}{accuracy} \\ \midrule
0     & 0.00 (0.00)                        & 1.29 (0.97)              & 1.00 (0.00)                  \\
1     & 1.98 (0.69)                        & 2.85 (2.49)              & 0.99 (0.10)                  \\
2     & 2.74 (1.07)                        & 5.06 (4.62)              & 0.98 (0.15)                  \\
3     & 3.17 (1.19)                        & 7.26 (5.78)              & 0.98 (0.13)                  \\
4     & 3.95 (1.84)                        & 11.59 (12.86)            & 1.00 (0.00)                  \\ \bottomrule
\end{tabular}%
}\caption{Mean (std) number of properties, response time, and accuracy of for directives with different maximum depth.}\label{tab:perceptual_study_stats}
\end{table}


For analysis, we subtract the length of the utterance audio from the response time and normalized the time for each subject independently. 




We estimate the cost function in Eq.~\ref{eq:cost}.
We set $\mathbf{D}_l = \begin{bmatrix}\mathbf{n}^0 & \mathbf{n}^1 & ... & \mathbf{n}^{d_{max}}\end{bmatrix}^{\mathbf{T}}$, where $\mathbf{n}^d =\begin{bmatrix}n_{p_1}^d & n_{p_2}^d & ...& n_{p_{max}}^{d} \end{bmatrix}$, each element representing number of certain types of property $p_i$ at depth $d$. Table~\ref{tab:cost_A2} shows the values of $\mathbf{C}_p^d$ found by linear regression. The intercept was 0.0386. The result suggests that certain properties like color, global location, size of the groups are more salient features than others. Interestingly, some properties like dialog- and task-context have contradicting effect for different depth levels.

\begin{table}[h]
\footnotesize \centering
\begin{tabular}{@{}lllllllll@{}}
\toprule
   & REL   & REF   & COL    & DM     & CON    & CNT    & ORD   & LOC    \\ \midrule
d0 & 0     & 0     & 0      & 0      & 0      & 0      & 0     & 0      \\
d1 & 0.099 & 0.021 & -0.053 & -0.026 & -0.042 & -0.051 & 0.077 & -0.107 \\
d2 & 0.088 & 0.064 & -0.065 & 0.055  & 0.114  & -0.068 & 0.183 & -0.145 \\
d3 & 0.096 & 0.032 & -0.116 & 0.096  & 0      & -0.084 & 0     & 0      \\
d4 & 0     & 0     & 0      & 0      & 0      & 0      & 0     & 0      \\ \bottomrule
\end{tabular}%
\caption{Cost of properties by their type at different depth.}
\label{tab:cost_A2}
\end{table}



\begin{figure}[h]
\centering
\includegraphics[width=0.6\textwidth]{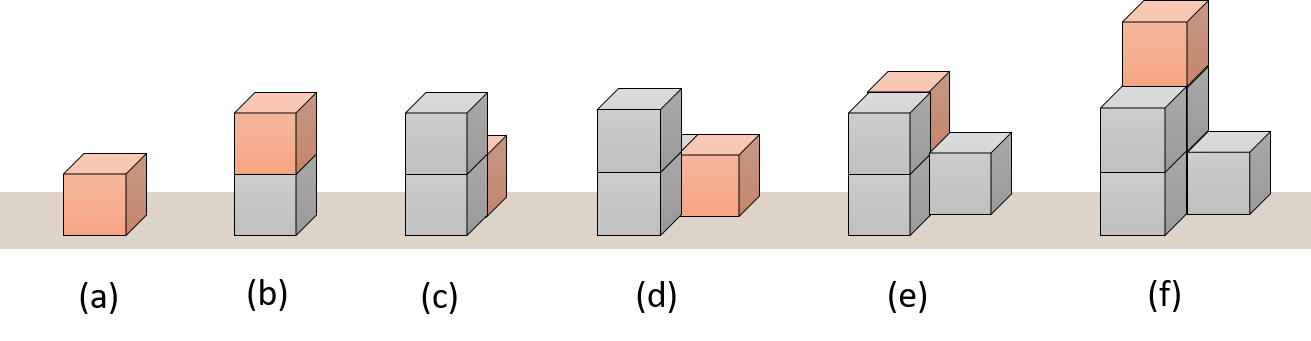}
\caption{Example of a block building plan used to generate results in Table~\ref{tab:plan_speech_result}. The highlighted block indicates the new block to be added.}
\label{fig:plan_example1}
\end{figure}

\paragraph{Results} Using the estimated weights in Table~\ref{tab:cost_A2} we can compute the cost of different alternative natural language options generated for a given directive. 
Table~\ref{tab:plan_speech_result} shows sets of directives generated by our algorithm for some steps of the plan shown in Figure~\ref{fig:plan_example1} and the corresponding cost values evaluated using the cost function. The red highlighted texts show the utterance choices adopted.

\begin{table}[h]\footnotesize
\centering
\begin{tabular}{@{}lllll@{}}
\toprule
  & Directive                                                                                                       & Depth & \#Prop & Cost  \\ \midrule
a & {\color[HTML]{FE0000}  Put a block on the table. }                                           & 1     & 2      & 0.158 \\
b & {\color[HTML]{FE0000}  Put a block on top of it. }                                           & 1     & 2      & 0.096 \\
  & \begin{tabular}[c]{@{}l@{}}Put a block on top of the block \\ you just placed.\end{tabular}  & 1     & 3      & 0.117 \\
c & {\color[HTML]{FE0000}  Put a block behind the stack.}                                        & 1     & 2      & 0.158 \\
  & Put a block behind the stack you just made.                          & 1     & 3      & 0.117 \\
  & Put a block behind the block at the bottom.                          & 2     & 3     & 0.246 \\
 d & {\color[HTML]{FE0000} Put a block to the right of it.}                                   & 1     & 2      & 0.096 \\
  & Put a block to the right of the previous block.                      & 1     & 3      & 0.117 \\
  & Put a block to the right of the block at the back.                   & 2     & 3      & 0.246 \\
e & {\color[HTML]{FE0000} Put a block behind the stack.}                                                            & 1     & 2      & 0.158 \\
  & {\color[HTML]{080808} Put a block behind the block on the top.}                                                 & 2     & 3      & 0.245 \\
  & \begin{tabular}[c]{@{}l@{}}Put a block on top of the block that is\\  at the back and on the left.\end{tabular} & 3     & 4      & 0.342 \\
f & {\color[HTML]{FE0000} Add one more.}                                                                            & 0     & 1      & 0.039 \\
  & Put a block on top of it.                                                                                       & 1     & 2      & 0.096 \\
  & Put a block on top of the same stack.                                                                           & 1     & 3      & 0.117 \\
  & Put a block on top of the stack at the back.                                                                    & 2     & 3      & 0.246 \\
  & \begin{tabular}[c]{@{}l@{}}Put a block on top of the block that is \\ on the top and at the back.\end{tabular}  & 3     & 4      & 0.342 \\ \bottomrule
\end{tabular}%
\caption{Directives generated for the plan in Figure \ref{fig:plan_example1}.}
\label{tab:plan_speech_result}
\end{table}

\subsection{User Evaluation}
We performed larger scale user evaluation study to compare our algorithm and a naive generator. The naive generator generates unoptimized descriptions using combination of the properties of a block. We denote the set of directives generated by EGT algorithm and the naive generator as $G_e$ and $G_n$, respectively.

Subjects performed the same tasks as those from the perceptual study but they were given much smaller set of directives. We collected data from 16 subjects. Numbers of data collected is 192 for $G_e$, and 80 for $G_n$. Accuracy was 97\% and 74\%, average time (std) taken to respond to the question were 20.21 (26.91) seconds and 40.6 (42.75) seconds, respectively. The results shows significant improvement in performance measured by time ($p < 0.05$ by Welch's t-test~\cite{welch-t-test-1947}).

%% file: 4_conclusion.tex
\vspace{-0.2in}
\section{Discussion and Future Work}
We presented a method to generate optimal natural language directive from machine generated block placement directives in the Blocks World. Here efficiency was evaluated from empirically determined cost functions. Our definition of efficiency is currently limited to human-agent communication in the blocks world, where we are expected to communicate with complete sentences.
However, human-human communication is multi-modal and often consists of series of spontaneous interactions which are incomplete, ambiguous~\cite{ambiguous-spatial-referring-expression-wallbridge-2019} or overspecified~\cite{overspecification-KOOLEN-2011,overspecification-fernandez-2019}. We would like to incorporate studies about human perception~\cite{logic-of-perception-rock-1983} and human-human interactions~\cite{jara-ettinger_rubio-fernandez_2020} to provide more natural interactions with virtual agents.  We also want to expand to vocabularies learned from real-world data~\cite{bisk2016-dataset}. 

We are currently working on extending this work for multi-modal interactions where the agent also communicates with co-speech gestures. We are particularly interested in formulating efficiency when these gestures are involved. Paradoxically, preliminary experiments indicate that redundancy in the gesture modality increases efficiency. 



